# GM Score: Incorporating inter-class and intra-class generator diversity, discriminability of disentangled representation, and sample fidelity for evaluating GANs

Harshvardhan GM[1], Aanchal Sahu[1], Mahendra Kumar Gourisaria[1]

[1] School of Computer Engineering, KIIT Deemed to be University, Bhubaneswar, India

## Abstract

While generative adversarial networks (GAN) are popular for their higher sample quality as opposed to other generative models like the variational autoencoders (VAE) and Boltzmann machines, they suffer from the same difficulty of the evaluation of generated samples. Various aspects must be kept in mind, such as the quality of generated samples, the diversity of classes (within a class and among classes), the use of disentangled latent spaces, agreement of said evaluation metric with human perception, etc. In this paper, we propose a new score, namely, GM Score, which takes into various factors such as sample quality, disentangled representation, intra-class and inter-class diversity, and other metrics such as precision, recall, and F1 score are employed for discriminability of latent space of deep belief network (DBN) and restricted Boltzmann machine (RBM). The evaluation is done for different GANs (GAN, DCGAN, BiGAN, CGAN, CoupledGAN, LSGAN, SGAN, WGAN, and WGAN Improved) trained on the benchmark MNIST dataset.

## 1. Introduction

Generative Adversarial Networks (GAN) are known for achieving high sample fidelity by implicit density estimation of a noise distribution [1, 2]. They are applied in various fields such as medical imaging [3], super resolution [4], facial recognition [5], etc. Unlike other generative models that maximize likelihood explicitly but have intractable likelihood functions, GANs involve two components – a discriminator $D$ and generator $G$ that compete in a minimax game that terminates at a saddle point, or the Nash equilibrium [6], where $G$ generates samples good enough to fool $D$ such that $D$ is no longer able to discriminate properly between genuine and fake samples. The generated data is often used to compare different generative models; the ideal generative model is one that correctly captures the whole training distribution to generate visually consistent and diverse samples. After the base GAN, many models have been proposed that augment the model in order to cover some of its limitations, and in this paper, we use a few such models for comparison. More specifically, we shall evaluate the performance of deep convolutional GAN (DCGAN) [7], bidirectional GAN (BiGAN) [8], conditional GAN (CGAN) [9], coupled GAN (CoupledGAN) [10], least squares GAN (LSGAN) [11], semi-supervised GAN (SGAN) [12], Wasserstein GAN (WGAN) [13], and Wasserstein GAN improved (WGAN Improved) [14].

In this paper, we propose a new evaluation metric for GANs, namely, the *GM Score*[1]. For the same, we use the benchmark MNIST dataset [19]. With the proposed metric, we try to overcome the shortcomings of certain other evaluation metrics existent in the literature (discussed in Section 3). The proposed metric makes use of vision-based classifier ensembles, unsupervised methods (restricted Boltzmann machine (RBM) [20] and deep belief network (DBN) [21] for feature extraction, and incorporates two diversity terms to output a score that indicates the generative model performance as a whole. The rest of the paper is organized as follows – 2. Motivation and contribution, 3. Related work, 4. Preliminary, 5. GM Score, 6. Results, 7. Discussion, and 8. Conclusion.

## 2. Motivation and contribution

Historically, the evaluation of deep generative models has been a difficult task due to several reasons. While implicit density estimation-based generative models like the deep Boltzmann machines [15] and variational autoencoders (VAE) [16] tend to maximize log-likelihood, giving us a measure of model performance, GANs tackle the maximization of likelihood in terms of training the objective for the discriminator [1]. Hence, without an explicit likelihood measure, one has to look at other parameters for evaluating GANs. Even models that employ likelihood maximization suffer from several limitations [17, 2], mainly because likelihood maximization does not always conform to human perception. If GANs are evaluated based on sample fidelity, it could be possible that a

---
[1] The code for implementation of all GANs and score calculations can be found at github.com/GM-git-dotcom/GM-Score

memory GAN may come up with the best quality of samples – even log-likelihood favours a memory GAN. A memory GAN generates samples that are very similar to the training data, and essentially only performs a copy operation – which is something we seek to avoid in the field of generative modelling. Another important aspect for the evaluation of GANs is the extent of mode drop or mode collapse. Mode collapse refers to the model learning only a part of the whole support of the training distribution, thus leading to generated samples (which may be within the same class) having a lack of visual diversity [18]. The more obvious component of GAN evaluation is sample fidelity, and various approaches have been employed in the past that try to favour models that generate data that best resembles the training distribution. Finally, it is also important to note that the diversity of generated samples, in terms of covering the entire class distribution, is also looked at as an important feature in a generative model. Often evaluation measures fail to incorporate both, the inter-class and intra-class diversities of the generated samples. We shall discuss some previously proposed GAN evaluation measures in Section 3. Combining all these objectives of evaluation in a single performance metric is a difficult task, hence, this motivated us to come up with a new metric for different GANs on a simple benchmark dataset to label the models with scores that represent a) inter-class generator diversity, b) intra-class generator diversity, and d) sample fidelity through discriminability of the samples when evaluated by unsupervised neural networks such as the DBN and RBM. To the best of our knowledge, no metric has been devised to incorporate all these elements. We enlist the main contributions of our paper as follows –

- We propose a novel metric for GAN evaluation (also applicable to generative models in general that are used for data generation).
- Various models are taken into consideration (GAN, DCGAN, BiGAN, CGAN, CoupledGAN, LSGAN, SGAN, WGAN, and WGAN Improved).
- As opposed to other approaches, the proposed metric takes both, the generator diversity (inter and intra-class), and sample fidelity as objectives for the output score.
- We obtain sample fidelity in a novel way – by combining discriminability of generated samples by using two different unsupervised neural networks, and convolutional neural network (CNN)-based classifier ensembles trained and tested (interchangeably) on original and generated data distributions.
- Unlike many other approaches, we explicitly find the intra-class diversity which is important to detect mode collapse and understand how many modes of the support are being captured by the generator.

## 3. Related work

GAN evaluation measures are generally classified into two subtypes: qualitative and quantitative. The former relies on the inspection of the quality of the samples generated, while the latter uses methods like minimizing the Kullback-Leibler (KL) divergence between data generated under the model and true data [22]. Inversely, it can be stated that maximizing the data likelihood is equivalent to minimizing the KL divergence. Likelihood density estimation is one of the most popular quantitative measures for the performance of GANs. In this paper, we mainly deal with quantitative analysis of the GANs on the MNIST dataset, hence, we shall look at only approaches that fall under this categorization.

One of the most widely accepted quantitative metrics for GAN evaluation is the Inception Score (IS), proposed by Salimans et al. [23] (2016). It makes use of transfer learning – with ImageNet [24] weights – using the InceptionNet [25] to classify generated images. Given the vectors of image labels $y$ and data $x$, the IS produces two distributions – a conditional label distribution $p(y|x)$ and a marginal distribution $\int p(y|x = G(z))dz$, where $G(z)$ refers to the generated image, $z$ is the noise distribution and $G$ is the GAN generator. The conditional label distribution is obtained by discriminating generated images into classes using the pre-trained InceptionNet, and the marginal distribution is simply the sum of all the conditional label distributions for each generated sample. The score calculates the KL divergence between these two distributions (conditional label distribution having low entropy, and marginal high) and favours models that have higher divergence – which means that images are highly discriminable and the generated distribution contains a good variety of samples from each class. We shall compare the proposed metric in Section 7, where we show how our metric takes into consideration more aspects of model evaluation.

Since IS does not consider intra-class diversity, a model trapped in a single mode may score high. Because of its lack of sensitivity to mode collapse, more research was required in capturing the intra-class diversity of generated data. Owing to this, the modified IS (m-IS) was proposed by Gurumurthy et al. [26] (2017) who replaced the KL term of IS with a cross-entropy, $p(y|x_i)\log\big(p(y|x_j)\big)$ where $x_i$ and $x_j$ are samples from the same class. Essentially, they calculate the KL divergence between each sample's conditional class distributions produced by the InceptionNet against that of another sample of the same class. This process may prove to be computationally expensive, as finding the divergence for $n$ samples would consume a complexity of $\mathcal{O}(n(n-1)) \approx \mathcal{O}(n^2)$ which is averaged per class. We solve this problem in a more simplified manner which consumes only $\mathcal{O}(n)$ complexity, which will be discussed in detail in Proof 6. Putting more emphasis on the importance of the prior training distribution which is ignored by the IS, Che et al. [27] (2016) introduced the mode score that finds the difference of KL divergence between label distribution and $p(y|x)$ and $p(y)$.

Heusel et al. [28] (2017) introduced the Fréchet Inception Distance (FID), that considers the embedding of generated samples into a feature space obtained through the activation of a layer in the InceptionNet. They consider this embedding as a continuous multivariate Gaussian, and estimate the first two moments, mean and covariance, of these Gaussians (of true and generated data). Then, the Fréchet Distance [29] (or the Wasserstein-2 distance [30]) is calculated between the two Gaussians to evaluate the performance of the generative model. While FID might be superior in terms of robustness to noise in comparison to IS, it assumes that the InceptionNet layer coding is a continuous multivariate Gaussian. The Generative Adversarial Metric (GAM), proposed by Im et al. [31] (2016), evaluates two GAN models by swapping their generators and discriminators. Given two GANs $\omega_1$ and $\omega_2$, with components $(D_1, G_1)$ and $(D_2, G_2)$ respectively (pertaining to discriminator $D$ and generator $G$), it is hypothesized that $\omega_1$ performs better than $\omega_2$ if $G_1$ is able to fool $D_2$ a higher number of times than the number of times $G_2$ is able to fool $D_1$. This is estimated by computing the likelihood ratio of the models $\omega_1$ and $\omega_2$. While this may be an intuitive way to compare different GANs, to "fool" the discriminator of another GAN implies that the ability of the discriminator is important. If the other discriminator is easy to fool, then the model in question wins easily, thus, both discriminators must be trained on a calibration dataset to have somewhat equal capability in discriminating images.

The GAN Quality Index (GQI), proposed by Ye et al. [32] (2018) makes use of a concept that is closely related to one of the subtasks to calculate the GM Score: the classifier ensembles, which we shall discuss in further sections. In the GQI, after training a GAN on data $x$ with labels $y$, a classifier is trained on the same dataset. It is tested on the generated images to obtain predicted class labels. Another classifier is then trained on generated data $\tilde{x}$ and both these classifiers are tested on the same real data to obtain a ratio of their test accuracies. This yields an integral value lying in [0, 100], wherein a higher GQI value indicates better generative performance. In our work, instead of a classifier, we use a deep ensemble to replicate a bagging approach to augment performance. Moreover, we do not calculate a ratio between accuracies, but instead plug the obtained accuracy with scores of other measures such as inter-class and intra-class diversity, and latent space discriminability.

## 4. Preliminary

In this section, we shall describe the sub-components that make up the computation or estimation of the GM Score. It involves four parts: i) calculating inter-class diversity, ii) calculating intra-class diversity, iii) latent space discriminability, and iv) classifier ensembles. We also describe the MNIST dataset along with the training and testing sets.

### 4.1 Inter-class generator diversity estimation

It is important for a generator to model the whole support of the dataset, which includes all the different classes it has seen during training. The ideal case, then, is when $G$ is asked to generate $n$ samples, it generates $n/C$ samples for each class $c_i$, where $C$ is the number of classes in the training dataset. In the GM Score, $\mathbb{D}^:$ is the diversity term (consider symbol : as vertical traversal through all classes) that measures the extent to which the generated samples $G(z)$, generated over latent noise vector $z$, properly cover all the classes $C$ in terms of the number of samples.

$$\mathbb{D}^:\big(G(z)\big) = \frac{\frac{\sum_{i=1}^{\|C\|}\|c_i\|}{\|C\|} - \frac{\sum_{i=1}^{\|C\|}\left(\left|\frac{\sum_{i=1}^{\|C\|}\|c_i\|}{\|C\|} - \|c_i\|\right|\right)}{\|C\|}}{\frac{\sum_{i=1}^{\|C\|}\|c_i\|}{\|C\|}} \equiv 1 - \frac{\text{avg. distance}\left(\|I_g\|, \mu(\|I_g\|)\right)}{\mu(\|I_g\|)} \qquad (1)$$

In (1), we first find the mean $\mu(\|I_g\|)$ (of the generated image distribution $I_g$) and subtract it with the average distance of the image counts in $I_g$ from the mean $\mu$. This difference is divided by $\mu(\|I_g\|)$ to obtain a diversity term that is of the bounds $\mathbb{D}^{:}(G(z)) \in (0, 1]$. Note that the operator $\|.\|$ refers to the number of generated samples in the distribution, hence, for instance, $\|C\|$ and $\|c_i\|$ refer to the total number of samples in all classes $C$ and in the $i^{th}$ class $c_i \in C$, respectively. The two extreme cases of the operation of generator $G(z)$ are discussed in Lemma 1 and Lemma 2. Lemma 3 validates our calculation for diversity to lie in the bounds of (0, 1].

**Lemma 1.** Inter-class diversity $\mathbb{D}^{:}(G(z)) = 1$ when images of all classes are generated equally.

**Proof 1.** Equation (1) effectively calculates how far each count of every class is from the mean, adds them together, and this sum is divided by the mean, which normalizes it to lie in the range (0, 1]. Thus, what this yields is an indicator of the diversity of generator $G(z)$, where if each class's sample count equals the mean of all classes (which means the generator is giving equal importance to each class), the average distance between each class's generated sample counts $\|I_g\|$ and the mean $\mu(\|I_g\|)$ becomes 0, and thus $\mathbb{D}^{:}(G(z)) \xrightarrow{yields} 1$. ∎

**Lemma 2.** $\mathbb{D}^{:}(G(z)) \approx 0$ when generator $G(z)$ is biased towards certain classes in $C$.

**Proof 2.** In the other extreme case when generator $G(z)$ does not generate samples equally, distribution $I_g$ will be non-uniform. With (1), we can calculate this non-uniformity by calculating the distance of each class count in $I_g$ from the mean $\mu(\|I_g\|)$. The division term on the RHS in (2) will always lie in the range $[0, 1)$, hence when subtracted from 1, the range becomes (0, 1]. This is proven in Lemma 3. So, the biased generator $G(z)$ will obtain a lower $\mathbb{D}^{:}$ since $\left(\text{avg. distance}/\mu\right) \approx 1$ as avg. distance approaches $\mu$. ∎

**Lemma 3.** $\mathbb{D}^{:}(G(z))$ strictly holds avg. distance $\left(\|I_g\|, \mu(\|I_g\|)\right) < \mu(\|I_g\|)$.

**Proof 3.** Essentially, the function avg. distance(a, b) is the mean absolute deviation (MAD) of a distribution, defined by $\frac{\sum|x_i - \mu|}{n}$, where $n$ is the number of data points. It is quite evident, that the mean, calculated by $\mu = \frac{\sum x_i}{n}$, is greater than $\frac{\sum|x_i - \mu|}{n}$. Inversely, MAD is always less than mean. Hence, Lemma 3 is proved. ∎

### 4.2 Intra-class generator diversity estimation

Intra-class diversity, denoted by $\mathbb{D}^{\cdots}$ (consider symbol $\cdots$ as horizontal traversal through samples of the same class), is usually more difficult to achieve for a generator. The rationale behind this is the limited capacity of the unimodal discriminator to discriminate multimodal data. It is easy for a metric to assign a high score to a generator that has dropped modes, a process which makes it easier for generators to generate only certain modes of the true data. In the proposed metric, we use a classifier trained on real data (MNIST), to output labels for each generated image. This gives us a conditional class label distribution $p(y|x)$ on data $x$, and we seek to maximize the entropy $\mathcal{H}(p(y|x))$, which provides the extent of variation of samples in the same class. Formally, in the case of a single class, $\mathbb{D}^{\cdots}(G(z))$ can be described by (2) as,

$$\mathcal{H}(p(y|x \in c_i)) = -\sum_{x=c_i^1}^{c_i^{\|c_i\|}} p(y|x \in c_i) \ln(p(y|x \in c_i)) \tag{2}$$

Average entropy $\mathcal{H}_{avg}(\cdot)$ can be found over all classes by (3) as,

$$\mathbb{D}^{\cdots}(G(z)) = \mathcal{H}_{avg}(p(y|x \in C)) = -\sum_{i=1}^{\|C\|} \sum_{x=c_i^1}^{c_i^{\|c_i\|}} p(y|x \in c_i) \ln(p(y|x \in c_i)) \div \|C\| \tag{3}$$

If $p(y|x)$ is a uniform distribution, this would yield $\mathbb{D}^{\cdots}(G(z)) \to \infty$ (a high entropy value). It is important to note that the bounds of $\mathcal{H}$ are $[0, \infty)$. We shall prove these bounds below.

**Lemma 4.** A sharp class conditional label distribution yields an intra-class diversity $\mathbb{D}^{\cdots}(G(z))$ of zero.

**Proof 4.** Consider conditional class label distribution to be a set of softmax probabilities, $p(y|x \in c_i) = \{p_1, p_2, \ldots, p_{||y||}\}$ where $||y||$ is the number of labels (equivalent to $||c_i||$), $p_j$ refers to the probability of label $y_j \in c_i$. A sharp class conditional label distribution would mean that exactly one label $p_j$ equals unity, and the rest zero (probabilities are averaged over all samples in class $c_i$). In such a case, the classifier has the maximum confidence that all samples are the label ($p_j = 1$), and thus $\mathcal{H}$ is calculable as,

$$\mathcal{H}(p(y|x \in c_i) = \{0, 0, \ldots, 1, \ldots, 0\}) = -\sum_{x=c_i^1}^{c_i^{||c_i||}} p(y|x \in c_i) \ln(p(y|x \in c_i))$$

$$\Rightarrow \mathcal{H}(\{0, 0, \ldots, 1, \ldots, 0\}) = -\left((0 \times \ln 10^{-6}) + (0 \times \ln 10^{-6}) + \cdots + (1 \times \ln 1) + \cdots + (0 \times \log_2 10^{-6})\right)$$

$$\Rightarrow \mathcal{H}(\{0, 0, \ldots, 1, \ldots, 0\}) = 0$$

Above, we substitute 0 for $10^{-6}$ so the $\ln$ calculation is possible. In reality, too, no classifier would output 0 probability for a prediction. Since $\mathbb{D}^{\cdots}(G(z))$ is just the entropy averaged over all classes, if generator $G$ lacks intra-class diversity, it is obvious, based on the proof above, that $\mathbb{D}^{\cdots}(G(z)) \xrightarrow{yields} 0$. ∎

**Lemma 5.** A uniform conditional class label distribution yields $\mathbb{D}^{\cdots}(G(z)) \in \mathbb{R}^+$, and increasing the number of classes makes $\mathbb{D}^{\cdots} \to \infty$.

**Proof 5.** Following Proof 4, consider that the softmax probabilities are now uniform, e.g. $\{0.1, 0.1, \ldots, 0.1\}$. We can calculate $\mathcal{H}$ now as below,

$$\mathcal{H}(p(y|x \in c_i) = \{0.1, 0.1, \ldots, 0.1\}) = -\sum_{x=c_i^1}^{c_i^{||c_i||}} p(y|x \in c_i) \ln(p(y|x \in c_i))$$

$$\Rightarrow \mathcal{H}(\{0.1, 0.1, \ldots, 0.1\}) = -\left((0.1 \times \ln 0.1) + (0.1 \times \ln 0.1) + \cdots + (0.1 \times \ln 0.1)\right)$$

As each softmax probability is 0.1, the number of labels $||y|| = 10$, then,

$$\Rightarrow \mathcal{H}(\{0.1, 0.1, \ldots, 0.1\}) = -||y|| \times (0.1 \times -2.30) = 2.30$$

$$\Rightarrow \mathcal{H}(\{0.1, 0.1, \ldots, 0.1\}) = 2.30 \in \mathbb{R}^+$$

It is evident that as we increase the number of labels, for a uniform distribution, $\mathcal{H}$ will approach $\infty$, may it be asymptotically. Hence, Lemma 5 is proved. ∎

**Lemma 6.** The computational upper-bound on calculating $\mathbb{D}^{\cdots}(G(z))$ is $\mathcal{O}(n)$ for $n$ generated samples.

**Proof 6.** It is straightforward to infer from (2) that for a given generated class $c_i$, each sample is visited only once for the computation of entropy. This means that the complexity of calculating entropy for a class having $n$ generated samples is of the order $\mathcal{O}(n)$, as opposed to m-IS, that has a complexity of $\mathcal{O}(n^2)$ due to the consideration of KL divergence between every pair of generated samples in a class. ∎

Maximizing entropy $\mathcal{H}(p(y|x))$ means that the conditional label distribution is more uniform (however, rarely purely uniform), meaning that the classifier's discriminability is reduced owing to higher intra-class variance of samples, which is desirable. The entropies are calculated for each class separately (by averaging over all images in the class), and then averaging to obtain one entropy score for each GAN's generated images. We show the results of class-wise entropy for each GAN, along with their standard deviations, in Table 4. It is possible to detect intra-class mode collapse with the distribution of $\mathbb{D}^{\cdots}(G(z))$. In the case when the generator collapses to a single mode, the distribution $\mathbb{D}^{\cdots}$ will have a very low standard deviation, as all the samples in the class are similar

(hence yielding similar entropy). This critical standard deviation $\sigma_{crit}$ in our experiments was found to be close to 0.2. It is important for a practitioner to see whether, for a given model and class, $\sigma(\mathbb{D}^{\cdots})$ is absurdly low, which is indicative of mode collapse to a single instance.

$$\sigma_{crit} = \sqrt{\frac{\sum\left(\mathcal{H}^{(i)} - \mathcal{H}_{avg}(p(y|x \in C))\right)^2}{||C||}} \sim 0.2 \qquad (4)$$

Deciding a value for $\sigma_{crit}$ is based on empirical observation and might be subjective since we do not numerically arrive at the value of 0.2. It is solely based on our observation of the standard deviations of entropy distributions and visual analysis of the generated images to correlate among the two. In other words, if the standard deviation of entropies for a GAN averaged over all classes is 0.4, and we find that a large enough sample of the generated images show a good amount of variety in the same class, we identify 0.4 as a good standard deviation to have. We also find that the standard deviation has a positive correlation with the average entropy obtained, which makes it easier for us to omit its inclusion in the calculation of the GM Score. From visual analysis, we find that no generator falls into the single-mode trap for any class, so we can surely say that the $\sigma$ obtained for each class does not fall below $\sigma_{crit}$, and we set a value for it as 0.2. For regularization, when a generator does fail and captures only a few modes (or even a single mode), one shall find $\sigma < \sigma_{crit}$, and then the only plausible regularization for that class's $\mathbb{D}^{\cdots}(G(z))$ would be to modify it to zero, i.e. $\mathbb{D}^{\cdots}(G(z)) \to 0$. If this is true for all classes, then intra-class diversity as a whole becomes zero.

In the case where the generator collapses to a few modes (as compared to many modes of the true data), it is best to plot the entropy distribution and identify different peaks, where each peak would refer to each mode of the training data. None of these cases arise in our evaluations.

**Remark 1.** In our experiments, it was noticed that WGAN had scored very high on intra-class diversity, mostly due to bad sample quality. It is obvious that the conditional class label distribution for a CNN trained on MNIST and tested on the generated samples will be more uniform, equivalent to the model being unsure of what digit is fed, if the generation quality is low. Due to the distribution uniformity, an unusually high entropy is achieved, perhaps wrongly. Hence, we define an overdiversity coefficient $\beta$ (which we set $\beta = 0.5$), where, if a model's $\mathbb{D}^{\cdots}(G(z))$ is greater than $\beta$, the intra-class diversity $\mathbb{D}^{\cdots}$ is modified as,

$$\mathbb{D}^{\cdots}(G(z)) \leftarrow \beta - |\mathbb{D}^{\cdots}(G(z)) - \beta|, \text{iff } \mathbb{D}^{\cdots}(G(z)) > \beta \qquad (5)$$

It can be said that $\beta$ is essentially a penalizing factor or regularization parameter that ensures that models do not exploit the fact that a higher entropy (or intra-class diversity) is obtained for samples of lower fidelity.

**Remark 2.** There is a caveat to this modification, however, we do not address it in this paper to retain simplicity. The problem with modifying $\mathbb{D}^{\cdots}$ by $\beta - |\mathbb{D}^{\cdots} - \beta|$ is that the outer difference may become negative, whenever $\mathbb{D}^{\cdots} > 2\beta$. In these cases, the intra-class entropy is extremely high because of a uniform predicted label distribution. For $n$ labels, the highest entropy is achieved when the distribution is purely uniform. The maximum entropy for $n$ label distributions varies in-proportion (see Table 1 for entropies calculated for softmax probabilities[2]). For a CNN, when every label has an equal probability, this means that either the image is an out-of-distribution example, or, the model itself is unable to discriminate well. Both these cases are extremely rare and practically improbable to ever happen in the evaluation of GANs, which is why we do not address this extreme case, in which the GM Score may output negative values. For simplicity, if ever the GM Score outputs negative values, it is because there is an abnormally high intra-class diversity. It is left to the practitioner to choose a suitable $\beta$, where, if a model clearly does not generate proper samples, its intra-class diversity always lies above $\beta$, which is also why we term it as the overdiversity coefficient.

**Table 1.** Increment of $\mathcal{H}$ as the number of labels increase in a uniform distribution.

| # labels | $\mathcal{H}$ |
|---|---|
| 10 | 2.3025 |

---

[2] By softmax probabilities, we refer to all the probabilities summing to unity.

|     |         |
|-----|---------|
| $10^2$ | 4.6051  |
| $10^3$ | 6.9077  |
| $10^5$ | 11.5129 |

### 4.3 Discriminability of latent feature space

Along with diversity, it is equally important that the sample fidelity of a model remains high. It is noticed that a high log-likelihood does not always ensure a high sample quality [2], hence, in the proposed metric, we instead extract meaningful features from two prominent generative models – the restricted Boltzmann machine (RBM), and the deep belief network (DBN). In both models, the extracted features are mapped onto a latent space, and a logistic classifier is used to discriminate the features into classes from 0 to 9. Being a fully-connected bipartite, RBM has two layers – hidden and visible. The visible layer $v$ consists of random variables (or nodes) whose states are dependent on their connected nodes. Just like in a normal neural network, the parameterization of the model is achieved through weights in the connections, along with a bias or intercept for each node. If we denote the weight between node $r$ and $s$ by $w_{rs}$, there exists an energy function $E(v, h)$ which the model tries to minimize for each state, given by (6), where $b$ and $c$ are biases for visible and hidden layers, respectively.

$$E(\boldsymbol{v}, \boldsymbol{h}) = -\sum_r \sum_s w_{rs} v_r h_s - \sum_r b_r v_r - \sum_s c_s h_s \quad (6)$$

Using (6), for the RBM, the joint probability $P(v, h)$ can be determined by plugging in $E(v, h)$ as,

$$P(\boldsymbol{v}, \boldsymbol{h}) = \frac{\exp(-E(\boldsymbol{v}, \boldsymbol{h}))}{\sum_{v,h} \exp(-E(\boldsymbol{v}, \boldsymbol{h}))} \quad (7)$$

The DBN, on the other hand, makes use of the RBM at the top level for training the bottom layers. It is not a two layer network like the RBM, and uses a greedy procedure [15] to infer layers one at a time. As the trained RBM is stacked on top of the other layers, what this results in is the elimination of an independent prior for the deepest layer to allow sampling from it unbiasedly. The feature extraction and classification process is shown by Fig. 1. A logistic classifier is used to discriminate features which gives us the output label for the generated image. It is necessary to mention that along with accuracy, we also calculate important evaluation metrics of class-wise precision, recall, and F1 score to diversify discriminability. These metrics are incorporated for the final score that is proposed.

### 4.4 Classifier ensembles

Consider two classifier ensembles given by $C = \{C_1, C_2, C_3, C_4, C_5\}$, and $C^r = \{C_1^r, C_2^r, C_3^r, C_4^r, C_5^r\}$, where $r$ denotes reversal of train and test domains. The normal classifiers in $C$ are trained on the MNIST dataset, while, by reverse, we refer to the classifiers in $C^r$ being trained on the generated samples. For the sake of experimentation, we consider five different convolutional neural networks (CNN) based on their architecture (for architecture details, see Section 6). Through the classifier ensembles, it becomes possible to calculate two important components of the proposed metric – a) discriminability of generated samples (i.e. accuracy), and b) intra-class diversity estimation through entropy of conditional label distribution given by a selected CNN architecture, which is defined in (3). For the accuracy, we use the classifier ensemble as a bagging-based model and get the label that is predicted by a majority of the classifiers $C_i \in C$ or $C^r$. The rationale behind choosing ensembles instead of individual classifiers is that deep ensembles are proven to be more effective in capturing different loss landscapes of a distribution [34] and also improving accuracy, essentially when the parameters are initialized randomly. To obtain class-wise entropies, we make use of $C_1$ (the choice of $C_1$ would largely be insignificant, because of the similar performance $\forall C_i \in C$), which is also used to obtain intra-class diversity. The bagging approach for obtaining classifier ensemble scores is described in Algorithm 1. The predictive function $\mathcal{F}(\mathbb{I}^{(J)}, C_i)$ returns a class label, denoting the output of classifier $C_i$ when fed image $\mathbb{I}^{(J)}$. In the input to Algorithm 1, we also provide true test labels $T_{label}$ that are used for calculating accuracy with the obtained (or predicted) labels $O_{label}$. Lastly, the function $\text{maxf}(\cdot)$ returns the most frequently occurring element from an array that is passed to it as input. We obtain the ensemble accuracy for both, $C$ and $C^r$ and return a score normalized in the bounds of [0, 1]. Suppose $\alpha_C$ and $\alpha_{C^r}$ denote accuracy of $C$ and $C^r$ respectively, then, we obtain a normalized ensemble score (ES) as,

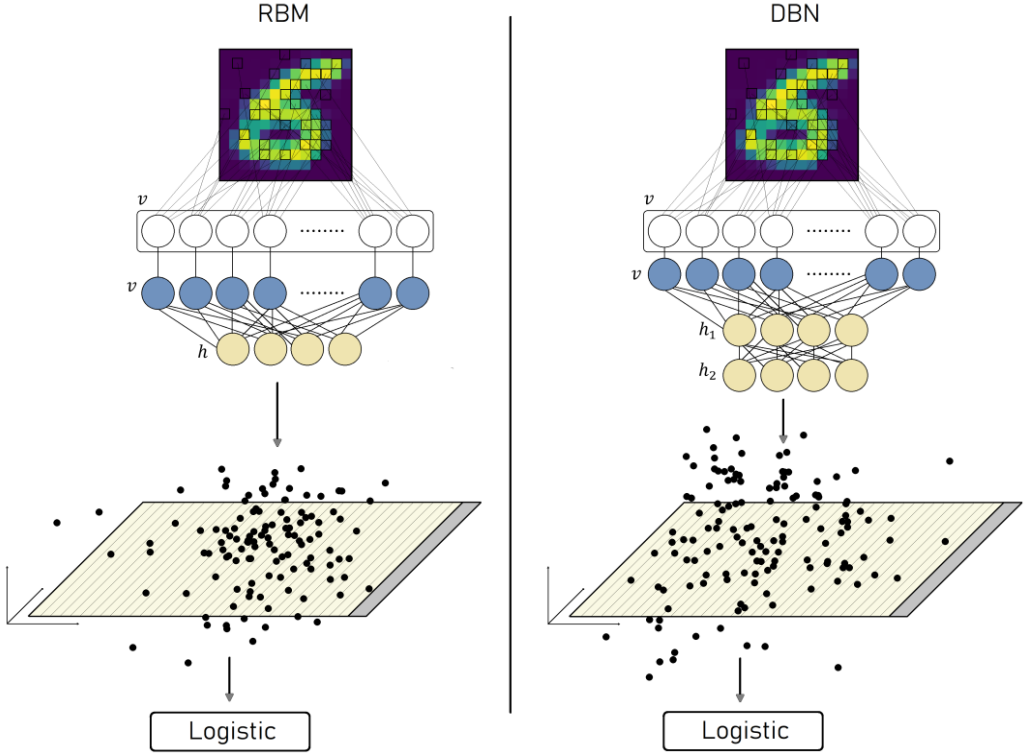

**Fig. 1.** L; Image discrimination process for RBM. R; Image discrimination process for DBN. Both models are trained on the MNIST (see training details in Section 6) training set and tested on samples generated by different GANs. This maps the features onto a latent space (this latent space is only a representation) which is discriminated by a logistic classifier [34]. Note that the hidden layer $h_1$ and visible layer $v$ are directed bottom-up, while all the other connections are bidirectional, even in the case of $v - h$ connections of RBM.

$$\text{ES} = \frac{100 - |\alpha_C - \alpha_{C^r}|}{100} \in [0, 1] \tag{8}$$

The value 100 in (8) is because the maximum accuracy possible is 100%. It is desirable for any GAN to obtain a higher ES score ($\approx 1$), which intuitively means that the performance of $C$ and $C^r$ are similar, i.e. the generated samples are good enough to replace true test samples of the MNIST dataset for learning the parameters of a CNN. The training and testing of these classifier ensembles is illustrated in Fig. 2.

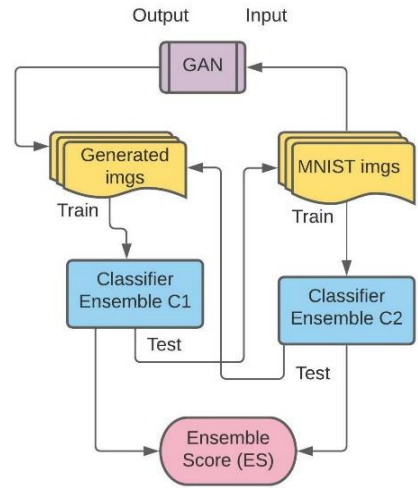

**Fig. 2.** Train-test domains are illustrated for $C$ and $C^r$ (depicted as C2 and C1, respectively).

**Algorithm 1.** Obtaining ensemble score ES for some GAN.

| **Input:** $\mathbb{I}_{test}$ ($m$ test samples), $C$, $C^r$ (for all GANs), $T_{label}$ | **Output:** ES |
|---|---|

1. $C \leftarrow \{C_1, C_2, C_3, C_4, C_5\}$, $C^r \leftarrow \{C_1^r, C_2^r, C_3^r, C_4^r, C_5^r\}$
2. $\mathbb{I}_{test} \leftarrow \{\mathbb{I}^{(1)}, \mathbb{I}^{(2)}, \mathbb{I}^{(3)}, \ldots, \mathbb{I}^{(m)}\}$
3. $O_{label} \leftarrow \{\}$
3. **for** ($j \leftarrow 1; j \leq m; j \leftarrow j + 1$): {
4.     $P \leftarrow \{\mathcal{F}(\mathbb{I}^{(j)}, C_1), \mathcal{F}(\mathbb{I}^{(j)}, C_2), \mathcal{F}(\mathbb{I}^{(j)}, C_3), \mathcal{F}(\mathbb{I}^{(j)}, C_4), \mathcal{F}(\mathbb{I}^{(j)}, C_5)\}$
5.     $O_{label}^{(j)} \leftarrow \text{maxf}(P)$ }
6. $s \leftarrow 0$
7. **for** ($j \leftarrow 1; j \leq m; j \leftarrow j + 1$): {
8.     **if** $T_{label}^{(j)} \equiv O_{label}^{(j)}$:
9.        $s \leftarrow c + 1$ }
10. $\alpha_c \leftarrow s/m$
11. **Repeat** steps 3 through 9 for $C^r$ pertaining to some GAN
12. $\alpha_{C^r} \leftarrow s/m$
13. **return** $(100 - |\alpha_c - \alpha_{C^r}| \div 100)$

### 4.5 MNIST

The original MNIST dataset [35] consists of a training set of 60,000 images and 10,000 images for the test set. The images are of dimensions 28×28 which are black and white. For training of classifier ensemble $C$, we consider 33602 images for training, and make use of image augmentation to enrich training data. The test set comprises 8398 images for testing (80:20 training to testing split). We generate 10,000 samples for each GAN to use the images as training data (whose labels are obtained using $C_1$, see Section 5 for more details) for classifiers in $C^r$, and then to be tested on a holdout test set comprising 2000 images. Digits in the MNIST dataset can be seen in Fig. 3. The training and testing class counts are given in Table 2.

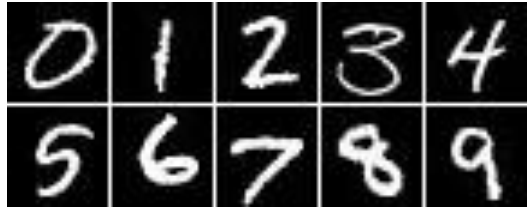

**Fig. 3.** A sample of digits in the MNIST dataset.

**Table 2.** Number of images in training and testing sets.

| Class | # of train images | # of test images |
|---|---|---|
| 0 | 3306 | 826 |
| 1 | 3748 | 936 |
| 2 | 3342 | 835 |
| 3 | 3481 | 870 |
| 4 | 3258 | 814 |
| 5 | 3036 | 759 |
| 6 | 3310 | 827 |
| 7 | 3521 | 880 |
| 8 | 3250 | 813 |
| 9 | 3350 | 838 |

### 5. GM Score

It is important to note that any generated image in $I_g$ does not have a corresponding label. Labels are important if we wish to train any $C^r$ on the generated domains. Initially, we took to the task of manually classifying (through human evaluation) 90000 images into respective classes, however, it proved to be very tedious. To circumvent this, we made use of $C_1$ to classify generated images $I_g$ and automated the process. We argue, that a classifier with over 99% accuracy on the original MNIST dataset, should be able to classify $I_g$ with equivalent proficiency to that of a human evaluator. For CGAN, since it generates class-conditional labels, this was not required to be done. For the rest of this paper, wherever necessary, we shall denote the GM Score with the symbol $\mathcal{G}(\cdot)$, where $\mathcal{G}$ is the GM Score function for a GAN. Consider that $\mathcal{D}^{F1}, \mathcal{D}^P, \mathcal{D}^R$ and $\mathcal{D}^\alpha$ refer to the DBN's F1-score, precision, recall and accuracy, which follows for RBM $\mathcal{R}$. When we combine all the components, $\mathcal{G}$ can be formulated as,

$$\mathcal{G}(\cdot) = \left[\frac{\mathcal{D}^{F1} + \cdots + \mathcal{D}^\alpha + \mathcal{R}^{F1} + \cdots + \mathcal{R}^\alpha}{8}\right] \times \left[\mathbb{D}^{\vdots}\big(G(\mathbf{z})\big)\right] \times \left[\frac{100 - |\alpha_c - \alpha_{C^r}|}{100}\right] \times \left[\mathbb{D}^{\cdots}\big(G(\mathbf{z})\big) - \beta\right] \quad (9)$$

In (9), $\beta$ is always zero unless $\mathbb{D}^{\cdots}\big(G(\mathbf{z})\big)$ exceeds 0.5 (see (5) for more details). We normalize $\mathcal{G}(\cdot)$ as follows, so it always lies in the range $[0, 1]$. Note that this normalization is similar to the one done in Section 4.4.

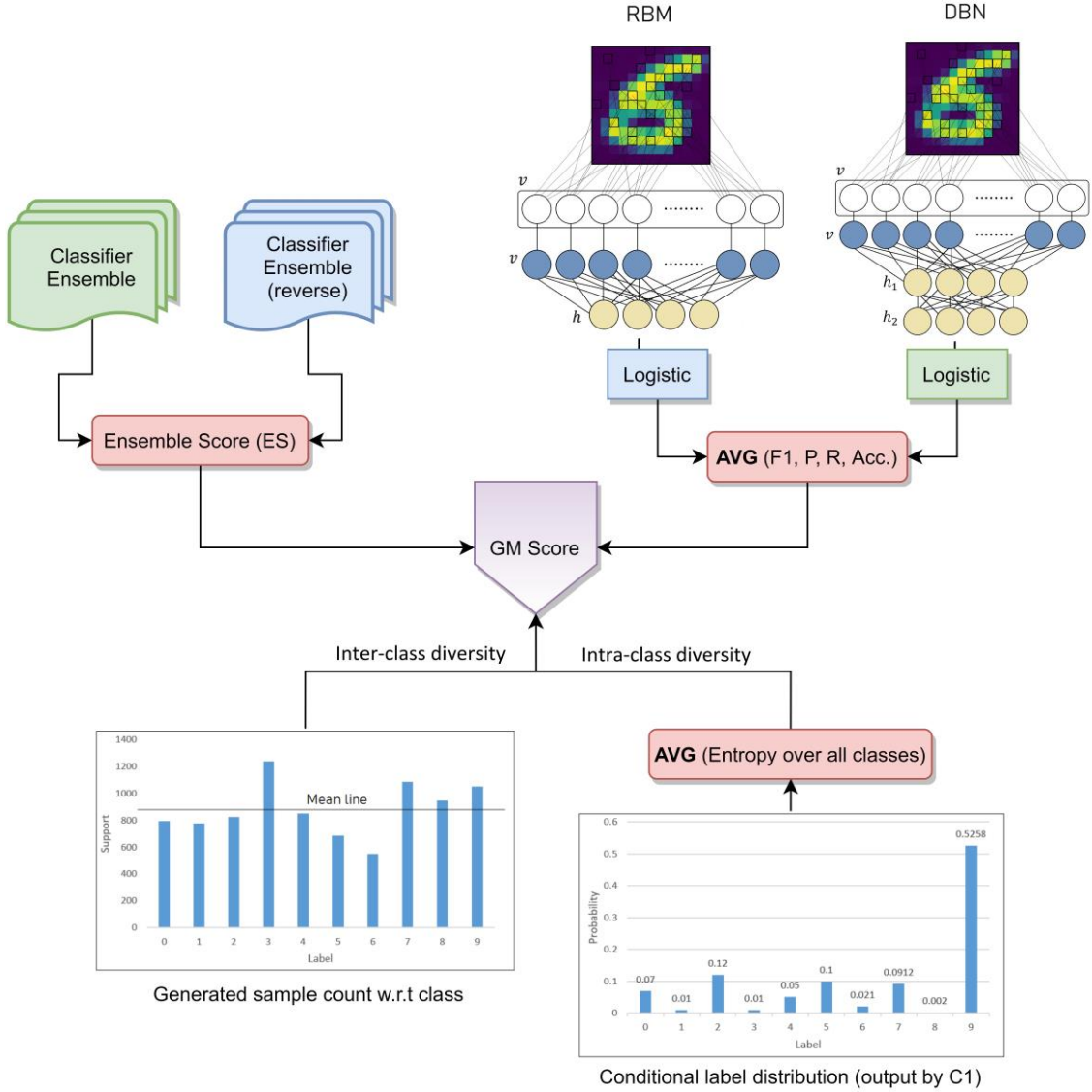

**Fig. 4.** High-level representative flowchart to calculate the GM Score.

$$\mathcal{G} \leftarrow \frac{\beta - |\beta - \mathcal{G}(\cdot)|}{\beta} = 1 - \frac{|\beta - \mathcal{G}(\cdot)|}{\beta} \quad (10)$$

The flowchart for obtaining $\mathcal{G}$ can be illustrated in Fig. 4.

**Lemma 7.** The lower and upper bounds of $\mathcal{G}(\cdot)$ are 0 and 1, respectively.

**Proof 7.** To prove this, we shall consider each individual component of the GM Score as formulated in (9). The metrics F1-score, precision, recall, and accuracy, output from the DBN and RBM lie in $[0, 1]$. Moreover, we know, from Lemma 1 and Lemma 2, that $\mathbb{D}^:(G(z)) \in (0, 1]$. From (9), we know that the ES term in (8), which is second-last from the left, lies in $[0, 1]$. Finally, the intra-class diversity $\mathbb{D}^{\cdots}(G(z))$, under normal conditions (see Remark 2), lies in $[0, 0.5]$. Hence, when we get a product $\mathcal{G}: [0,1] \times (0,1] \times [0,1] \times [0, 0.5] \rightarrow [0, 0.5]$, we normalize it as done in (10), to get $\mathcal{G}: norm([0,1] \times (0,1] \times [0,1] \times [0, 0.5]) \rightarrow [0, 1]$. ∎

**Remark 3.** The GM Score for CGAN is biased because inter-class diversity $\mathbb{D}^:(G(z))$ is unity. This is because all the image counts are equal, and thus the average distance from mean is zero (Lemma 1). The inter-class unity favours CGAN, however, the effects are not drastic, because other models also score $\mathbb{D}^:(G(z))$ close to 0.9. In general, the intra-class diversity component cannot be used for application to generative models that have a prior conditional distribution to sample from to obtain class-specific samples. This is because the model knows which class to generate beforehand, and evaluating how well it can generate diverse samples is illogical. Moreover, the latent space discriminability scores obtained from DBN and RBM also tend to be slightly higher, even though the image fidelity might be low. This is because CGAN generates modes from data that it receives in the conditional distribution. Once these features are detected by the RBM or DBN, it becomes easy for classification. Still, the score assigned to CGAN is not the best; the proposed score overcomes these problems.

## 6. Experiments and Results

In this section, we shall show the results of training and testing of models, along with architectural details, generated samples, and intermediate scores obtained to calculate $\mathcal{G}$. First, we shall calculate $\mathbb{D}^:(G(z))$ for every GAN. To reiterate, we have taken the following GANs under consideration: WGAN [13], WGAN Improved [14] (or WGAN_I), SGAN [12], LSGAN [11], GAN [1], DCGAN [7], CoupledGAN [10] (or CoGAN), CGAN [9], BiGAN [8]. Each GAN is trained at least for 20k (with some over 20k as they required more training) epochs to ensure convergence. Table 3 shows the image counts $||c_i||$ $\forall c_i \in C$, for every GAN. From Remark 3, it is evident why CGAN achieves the highest inter-class score possible, i.e. unity. Because CGAN generates images depending upon an input to sample from, the $||c_i||$ distribution is uniform.

**Table 3.** Inter-class diversity for each GAN on MNIST data generation.

| | Generated class image count $||c_i||$ | | | | | | | | | | |
|---|---|---|---|---|---|---|---|---|---|---|---|
| | 0 | 1 | 2 | 3 | 4 | 5 | 6 | 7 | 8 | 9 | $\mathbb{D}^:(G(z))$ |
| WGAN | 964 | 448 | 1426 | 1678 | 639 | 919 | 693 | 849 | 1234 | 1150 | **0.7024** |
| WGAN_I | 873 | 1016 | 1120 | 1185 | 858 | 826 | 1093 | 910 | 1117 | 1002 | **0.7942** |
| SGAN | 1025 | 1009 | 1046 | 913 | 876 | 775 | 1098 | 945 | 1218 | 1095 | **0.9017** |
| LSGAN | 832 | 1157 | 654 | 1535 | 740 | 963 | 946 | 1133 | 1004 | 1036 | **0.8270** |
| GAN | 952 | 996 | 991 | 1280 | 788 | 623 | 710 | 1130 | 1199 | 1331 | **0.8120** |
| DCGAN | 948 | 1043 | 925 | 1012 | 1019 | 794 | 1112 | 1008 | 1070 | 1069 | **0.9334** |
| CoGAN | 905 | 1120 | 851 | 730 | 891 | 758 | 589 | 1977 | 1144 | 1035 | **0.7448** |
| CGAN | 1000 | 1000 | 1000 | 1000 | 1000 | 1000 | 1000 | 1000 | 1000 | 1000 | **1.0000** |
| BiGAN | 792 | 777 | 826 | 1237 | 850 | 685 | 549 | 1084 | 947 | 1051 | **0.8182** |

We further visualize randomly sampled digits generated by each model in Fig. 5. It can be inferred that each GAN has a characteristic way of generating the images, for example, SGAN and DCGAN usually generate samples that are smoother, whereas CoGAN, CGAN, and BiGAN generate samples with random areas of activated pixels. It is also important to note that the visualization of generated samples in Fig. 5. are chosen for demonstrative purposes only, hence, they are not reflective of the actual average generation fidelity of the GANs. Next, we calculate the intra-class diversity $\mathbb{D}^{\cdots}(G(z))$ for each GAN by making $C_1$ predict class labels $\forall I_g$. This is obtained by the left-hand-side expression in (3), $\mathcal{H}_{avg}(p(y|x \in C))$, which is averaged over all classes. For each class, too, the entropies are averaged over all samples $I_g \in c_i$. The mean entropies, along with their standard deviations, are provided in Table 4. While it is clear that $\mathcal{H}_{avg} > 0.24$ for WGAN, the intra-class diversity is 0.24 because of the penalizing factor $\beta$. Other than WGAN, no other model obtained $\mathcal{H}_{avg} > \beta (= 0.5)$, hence, there is no regularization applied. This is also discussed in Remark 1. A higher intra-class diversity indicates the model's tendency to capture more modes of the data. Also, since no standard deviation is below $\sigma_{crit}$, it is clear that no model is collapsed to a single mode. This is also verified by visual observation of the samples as well, where if the generated samples in the same class were very similar, it would be visible.

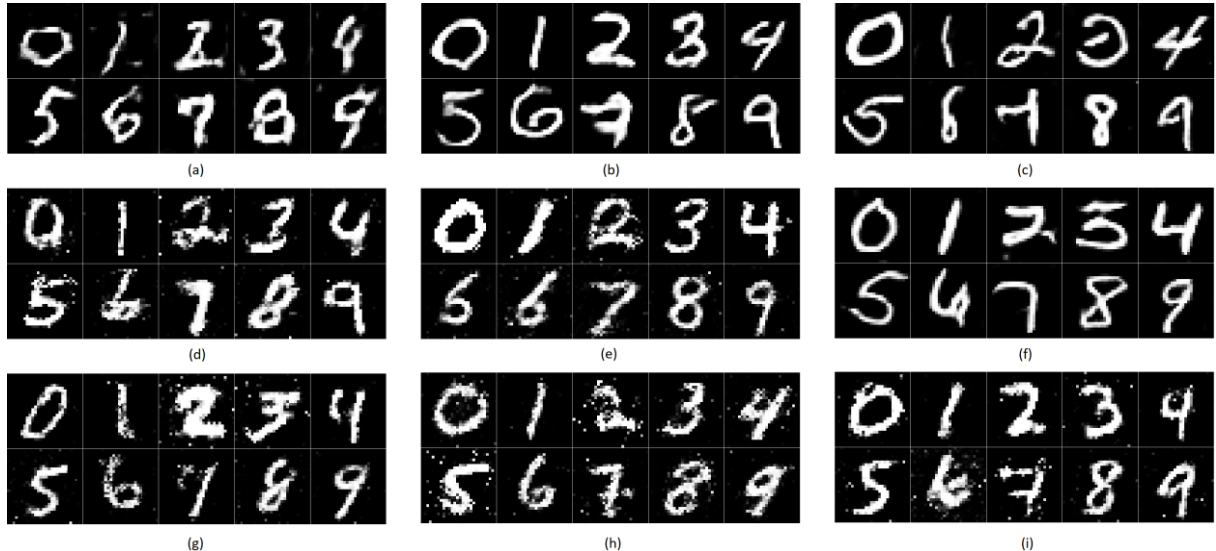

**Fig. 5.** Images generated from GANs (a) WGAN (b) WGAN_I (c) SGAN (d) LSGAN (e) GAN (f) DCGAN (g) CoGAN (h) CGAN and (i) BiGAN. Note: These images are random samples from the generated pool.

Table 4. Class-wise entropies averaged over $I_g$.

| $c_i$ | WGAN | WGAN_I | SGAN | LSGAN | GAN | DCGAN | CoGAN | CGAN | BiGAN |
|---|---|---|---|---|---|---|---|---|---|
| 0 | 0.58 ± 0.63 | 0.22 ± 0.44 | 0.21 ± 0.42 | 0.22 ± 0.44 | 0.19 ± 0.41 | 0.23 ± 0.44 | 0.22 ± 0.45 | 0.05 ± 0.22 | 0.28 ± 0.49 |
| 1 | 0.68 ± 0.63 | 0.15 ± 0.36 | 0.15 ± 0.38 | 0.13 ± 0.35 | 0.16 ± 0.39 | 0.14 ± 0.35 | 0.15 ± 0.39 | 0.09 ± 0.27 | 0.20 ± 0.41 |
| 2 | 0.92 ± 0.63 | 0.42 ± 0.53 | 0.44 ± 0.55 | 0.55 ± 0.58 | 0.46 ± 0.57 | 0.41 ± 0.54 | 0.51 ± 0.59 | 0.12 ± 0.30 | 0.65 ± 0.62 |
| 3 | 0.65 ± 0.60 | 0.35 ± 0.52 | 0.31 ± 0.47 | 0.27 ± 0.46 | 0.28 ± 0.47 | 0.33 ± 0.50 | 0.34 ± 0.50 | 0.09 ± 0.26 | 0.39 ± 0.54 |
| 4 | 0.70 ± 0.63 | 0.25 ± 0.45 | 0.19 ± 0.36 | 0.26 ± 0.46 | 0.25 ± 0.44 | 0.19 ± 0.37 | 0.27 ± 0.48 | 0.17 ± 0.37 | 0.37 ± 0.51 |
| 5 | 0.78 ± 0.61 | 0.29 ± 0.45 | 0.26 ± 0.44 | 0.25 ± 0.42 | 0.27 ± 0.44 | 0.25 ± 0.41 | 0.30 ± 0.48 | 0.26 ± 0.47 | 0.40 ± 0.52 |
| 6 | 0.97 ± 0.66 | 0.37 ± 0.55 | 0.33 ± 0.51 | 0.38 ± 0.55 | 0.32 ± 0.51 | 0.31 ± 0.51 | 0.39 ± 0.55 | 0.17 ± 0.39 | 0.60 ± 0.66 |
| 7 | 0.61 ± 0.60 | 0.18 ± 0.35 | 0.21 ± 0.41 | 0.16 ± 0.34 | 0.14 ± 0.31 | 0.21 ± 0.39 | 0.19 ± 0.37 | 0.16 ± 0.35 | 0.24 ± 0.41 |
| 8 | 0.87 ± 0.63 | 0.45 ± 0.55 | 0.37 ± 0.54 | 0.46 ± 0.58 | 0.37 ± 0.53 | 0.34 ± 0.51 | 0.44 ± 0.37 | 0.11 ± 0.28 | 0.64 ± 0.62 |
| 9 | 0.75 ± 0.59 | 0.35 ± 0.47 | 0.33 ± 0.47 | 0.37 ± 0.49 | 0.27 ± 0.42 | 0.33 ± 0.48 | 0.33 ± 0.45 | 0.22 ± 0.38 | 0.44 ± 0.50 |
| $\mathbb{D}^{\cdots}$ | **0.24** | **0.30** | **0.28** | **0.31** | **0.27** | **0.27** | **0.31** | **0.15** | **0.42** |

Now, we show the training results of the DBN and RBM on MNIST. We use these models as unsupervised feature extractors (Fig. 1) which are initially trained on the original dataset, and then tested on the generated samples for each GAN. The testing is done by using a logistic regression classifier that works on the appropriate features extracted by the unsupervised neural networks. The hyperparameters for DBN, RBM and the logistic regression classifier are given in Table 5, Table 6, and Table 7, respectively. The time complexity of implementing the RBM is $\mathcal{O}(n^2)$, where $n$ is the number of components or nodes in the hidden layer $h$. We also plot the DBN's pre-training reconstruction error for 20 epochs, where each hidden layer is trained for 10 epochs each, along with the pseudo-likelihood for the RBM over 10 iterations in Fig. 6. The features extracted by the RBM can be visualized in Fig. 7, which helps understand how these techniques are able to extract meaningful features to augment the performance of the logistic regression classifier. To see how the performance of the logistic regression classifier is enhanced with unsupervised features from RBM and DBN, the reader is advised to view the code results mentioned as a footnote in Section 1.

**Table 5.** Hyperparameters for DBN training. SGD refers to stochastic gradient descent [36], and $h$ structure is the hidden layer structure, with 256 nodes in the 1st layer and 512 in the 2nd. Finally, CD refers to contrastive divergence [37].

| $h$ structure | Batch size | RBM learning rate | Activation | CD iterations | Optimizer |
|---|---|---|---|---|---|
| [256, 512] | 10 | 0.06 | Sigmoid | 1 | SGD |

**Table 6.** Hyperparameters for RBM training. # components refers to the number of binary hidden units, or alternatively, $h$. PCD is persistent contrastive divergence used for the estimation of parameters [38].

| # components | Batch size | Learning rate | Random state | # iterations | Param estimation |
|---|---|---|---|---|---|
| 100 | 10 | 0.06 | 0 | 10 | PCD |

**Table 7.** Hyperparameters for logistic regression training, based on scikit-learn's implementation [39].

| Hyperparam | Value |
|---|---|
| **C** | 6000.0 |
| **Class weights** | None |
| **Dual** | False |
| **Fit intercept** | True |
| **Intercept scaling** | 1 |
| **L1 ratio** | None |
| **Max iterations** | 100 |
| **Multi-class** | Auto |
| **Number of jobs** | None |
| **Penalty** | L2 |
| **Random state** | None |
| **Solver** | Newton-cg |
| **Tolerance** | $10^{-4}$ |
| **Warm start** | False |

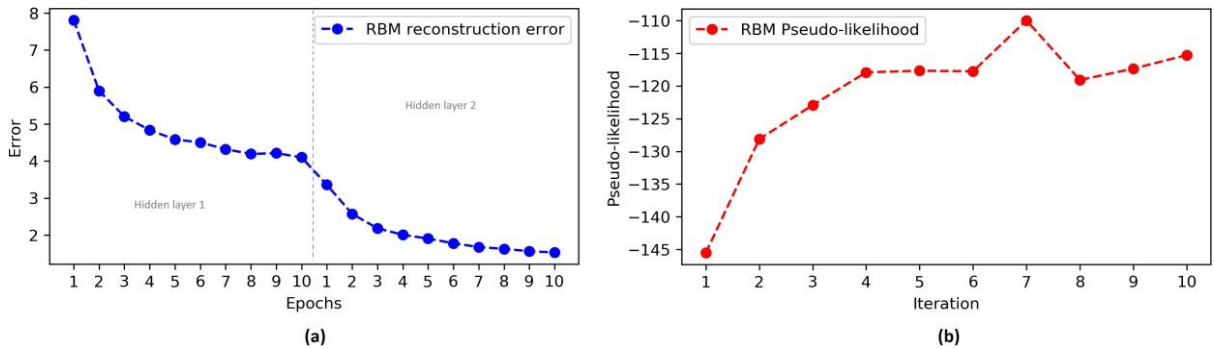

**Fig. 6.** (a) RBM reconstruction error for greedy DBN training procedure involving training each hidden layer separately for 10 epochs. This error is called "RBM reconstruction error" because for training each layer, the resultant model only has a visible and hidden layer i.e. equivalent to an RBM. (b) RBM pseudo-likelihood minimization with each iteration.

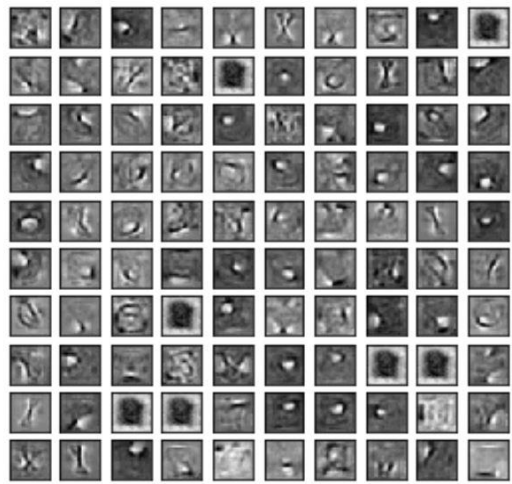

**Fig. 7.** Features extracted using RBM.

As mentioned in Section 4.3, along with accuracy, we also calculate evaluation metrics of precision, recall, and F1-score. Precision is defined as $TP/(TP + FP)$, recall as $TP/(TP + FN)$, and F1-score as $2 \times$ precision $\times$ recall/(precision + recall), where TP, TN, FP, FN refer to true positive, true negative, false positive, and false negative, respectively. These metrics are listed class-wise for WGAN, WGAN_I, SGAN, LSGAN, GAN, DCGAN, CoGAN, CGAN, BiGAN in Table 8, Table 9, and Table 10, respectively for DBN, and in Table 11, Table 12, and Table 13, respectively for RBM. In these tables, the values in bold are scores resulting because of low discriminability (or fidelity) of the generated samples. The explanation for why CGAN, which achieves a high score in both the cases of DBN and RBM as feature extractor, is given in Remark 3. It is noticed that even with CGAN samples that have a low fidelity (or noisy), the DBN and RBM are able to extract correct features for classification, the reason for which is given in Remark 3.

**Table 8.** Class-wise precision, recall, F1-scores, and accuracy for WGAN, WGAN_I and SGAN using DBN as the feature extractor.

|  | WGAN | | | WGAN_I | | | SGAN | | |
|---|---|---|---|---|---|---|---|---|---|
| $c_i$ | Prec. | Recall | F1 | Prec. | Recall | F1 | Prec. | Recall | F1 |
| 0 | 0.75 | 0.74 | 0.75 | 0.89 | 0.90 | 0.89 | 0.89 | 0.90 | 0.90 |
| 1 | 0.54 | 0.79 | 0.64 | 0.84 | 0.97 | 0.90 | 0.84 | 0.97 | 0.90 |
| 2 | 0.63 | 0.52 | 0.57 | 0.80 | 0.76 | 0.78 | 0.82 | 0.77 | 0.79 |
| 3 | 0.63 | 0.45 | 0.53 | 0.79 | 0.65 | 0.72 | 0.75 | 0.73 | 0.74 |
| 4 | 0.48 | 0.71 | 0.57 | 0.81 | 0.88 | 0.85 | 0.83 | 0.91 | 0.87 |
| 5 | **0.41** | **0.45** | **0.43** | **0.63** | **0.71** | **0.67** | **0.68** | **0.76** | **0.72** |
| 6 | 0.54 | 0.63 | 0.58 | 0.88 | 0.84 | 0.86 | 0.89 | 0.83 | 0.86 |
| 7 | 0.53 | 0.75 | 0.62 | 0.79 | 0.91 | 0.84 | 0.82 | 0.88 | 0.85 |
| 8 | 0.58 | 0.44 | 0.50 | 0.80 | 0.73 | 0.76 | 0.85 | 0.74 | 0.79 |
| 9 | 0.57 | 0.54 | 0.55 | 0.80 | 0.74 | 0.77 | 0.82 | 0.76 | 0.79 |
| Avg. | **0.57** | **0.60** | **0.57** | **0.80** | **0.81** | **0.80** | **0.82** | **0.83** | **0.82** |
| Acc. | **0.57** | | | **0.80** | | | **0.82** | | |

**Table 9.** Class-wise precision, recall, F1-scores, and accuracy for LSGAN, GAN and DCGAN using DBN as the feature extractor.

|  | LSGAN | | | GAN | | | DCGAN | | |
|---|---|---|---|---|---|---|---|---|---|
| $c_i$ | Prec. | Recall | F1 | Prec. | Recall | F1 | Prec. | Recall | F1 |
| 0 | 0.92 | 0.90 | 0.91 | 0.92 | 0.92 | 0.92 | 0.89 | 0.81 | 0.85 |
| 1 | 0.83 | 0.97 | 0.90 | 0.86 | 0.97 | 0.91 | 0.76 | 0.95 | 0.85 |
| 2 | 0.77 | 0.66 | 0.71 | 0.81 | 0.77 | 0.79 | 0.74 | 0.69 | 0.71 |
| 3 | 0.82 | 0.76 | 0.79 | 0.85 | 0.77 | 0.81 | 0.60 | 0.60 | 0.60 |
| 4 | 0.83 | 0.90 | 0.87 | 0.81 | 0.89 | 0.85 | 0.78 | 0.86 | 0.82 |
| 5 | 0.71 | 0.77 | 0.74 | 0.68 | 0.77 | 0.72 | **0.52** | **0.60** | **0.55** |
| 6 | 0.88 | 0.87 | 0.88 | 0.86 | 0.86 | 0.86 | 0.83 | 0.78 | 0.80 |
| 7 | 0.83 | 0.93 | 0.87 | 0.83 | 0.93 | 0.87 | 0.69 | 0.85 | 0.76 |
| 8 | 0.84 | 0.68 | 0.75 | 0.90 | 0.74 | 0.82 | 0.79 | 0.48 | 0.60 |
| 9 | 0.82 | 0.80 | 0.81 | 0.86 | 0.83 | 0.85 | 0.73 | 0.68 | 0.70 |
| Avg. | **0.83** | **0.82** | **0.82** | **0.84** | **0.85** | **0.84** | **0.73** | **0.73** | **0.72** |
| Acc. | **0.83** | | | **0.84** | | | **0.73** | | |

**Table 10.** Class-wise precision, recall, F1-scores, and accuracy for CoGAN, CGAN and BiGAN using DBN as the feature extractor.

|  | CoGAN | | | CGAN | | | BiGAN | | |
|---|---|---|---|---|---|---|---|---|---|
| $c_i$ | Prec. | Recall | F1 | Prec. | Recall | F1 | Prec. | Recall | F1 |
| 0 | 0.94 | 0.87 | 0.90 | 0.97 | 0.96 | 0.96 | 0.84 | 0.79 | 0.82 |
| 1 | 0.82 | 0.98 | 0.89 | 0.99 | 0.96 | 0.97 | 0.76 | 0.96 | 0.85 |
| 2 | 0.79 | 0.71 | 0.75 | 0.97 | 0.90 | 0.93 | 0.64 | 0.67 | 0.65 |
| 3 | 0.71 | 0.75 | 0.73 | 0.87 | 0.91 | 0.89 | **0.58** | **0.48** | **0.52** |
| 4 | 0.81 | 0.88 | 0.85 | 0.91 | 0.93 | 0.92 | 0.76 | 0.71 | 0.74 |
| 5 | 0.76 | 0.74 | 0.75 | 0.85 | 0.85 | 0.85 | **0.37** | **0.64** | **0.47** |
| 6 | 0.86 | 0.85 | 0.85 | 0.92 | 0.94 | 0.93 | 0.77 | 0.63 | 0.69 |
| 7 | 0.86 | 0.91 | 0.88 | 0.93 | 0.94 | 0.94 | 0.68 | 0.88 | 0.77 |
| 8 | 0.89 | 0.67 | 0.76 | 0.88 | 0.96 | 0.92 | 0.76 | 0.33 | 0.46 |
| 9 | 0.79 | 0.81 | 0.80 | 0.94 | 0.86 | 0.90 | 0.71 | 0.63 | 0.67 |
| Avg. | **0.82** | **0.82** | **0.82** | **0.92** | **0.92** | **0.92** | **0.69** | **0.67** | **0.66** |
| Acc. | **0.83** | | | **0.92** | | | **0.66** | | |

**Table 11.** Class-wise precision, recall, F1-scores, and accuracy for WGAN, WGAN_I and SGAN using RBM as the feature extractor.

|       | WGAN  |        |      | WGAN_I |        |      | SGAN  |        |      |
|-------|-------|--------|------|--------|--------|------|-------|--------|------|
| $c_i$ | Prec. | Recall | F1   | Prec.  | Recall | F1   | Prec. | Recall | F1   |
| 0     | 0.68  | 0.74   | 0.71 | 0.83   | 0.86   | 0.85 | 0.83  | 0.87   | 0.85 |
| 1     | 0.49  | 0.73   | 0.59 | 0.81   | 0.95   | 0.87 | 0.81  | 0.95   | 0.88 |
| 2     | 0.57  | 0.49   | 0.53 | 0.76   | 0.71   | 0.73 | 0.77  | 0.71   | 0.74 |
| 3     | 0.55  | 0.36   | 0.43 | 0.69   | 0.63   | 0.66 | 0.61  | 0.62   | 0.62 |
| 4     | 0.35  | 0.57   | 0.43 | 0.74   | 0.82   | 0.78 | 0.77  | 0.79   | 0.78 |
| 5     | 0.33  | 0.39   | 0.36 | **0.56** | **0.53** | **0.55** | 0.55 | 0.57 | 0.56 |
| 6     | 0.51  | 0.51   | 0.51 | 0.82   | 0.79   | 0.80 | 0.84  | 0.79   | 0.81 |
| 7     | 0.54  | 0.58   | 0.56 | 0.76   | 0.83   | 0.79 | 0.76  | 0.83   | 0.80 |
| 8     | 0.40  | 0.49   | 0.44 | 0.68   | 0.61   | 0.64 | 0.73  | 0.64   | 0.68 |
| 9     | 0.53  | 0.34   | 0.41 | 0.71   | 0.69   | 0.70 | 0.74  | 0.68   | 0.71 |
| Avg.  | **0.50** | **0.52** | **0.50** | **0.74** | **0.74** | **0.74** | **0.74** | **0.74** | **0.74** |
| Acc.  | 0.49  |        |      | 0.74   |        |      | 0.75  |        |      |

**Table 12.** Class-wise precision, recall, F1-scores, and accuracy for LSGAN, GAN and DCGAN using RBM as the feature extractor.

|       | LSGAN |        |      | GAN   |        |      | DCGAN |        |      |
|-------|-------|--------|------|-------|--------|------|-------|--------|------|
| $c_i$ | Prec. | Recall | F1   | Prec. | Recall | F1   | Prec. | Recall | F1   |
| 0     | 0.86  | 0.87   | 0.86 | 0.89  | 0.90   | 0.90 | 0.84  | 0.87   | 0.86 |
| 1     | 0.86  | 0.96   | 0.90 | 0.87  | 0.94   | 0.90 | 0.83  | 0.94   | 0.88 |
| 2     | 0.69  | 0.69   | 0.69 | 0.75  | 0.69   | 0.72 | 0.76  | 0.74   | 0.75 |
| 3     | 0.72  | 0.67   | 0.70 | 0.72  | 0.64   | 0.68 | 0.65  | 0.64   | 0.64 |
| 4     | 0.75  | 0.79   | 0.77 | 0.74  | 0.84   | 0.79 | 0.77  | 0.85   | 0.81 |
| 5     | **0.62** | **0.50** | **0.55** | **0.59** | **0.48** | **0.53** | **0.60** | **0.53** | **0.56** |
| 6     | 0.84  | 0.82   | 0.83 | 0.80  | 0.81   | 0.81 | 0.83  | 0.79   | 0.82 |
| 7     | 0.85  | 0.86   | 0.85 | 0.84  | 0.88   | 0.86 | 0.79  | 0.85   | 0.82 |
| 8     | 0.65  | 0.68   | 0.66 | 0.66  | 0.76   | 0.71 | 0.74  | 0.68   | 0.71 |
| 9     | 0.72  | 0.78   | 0.75 | 0.82  | 0.76   | 0.79 | 0.75  | 0.69   | 0.72 |
| Avg.  | **0.76** | **0.76** | **0.76** | **0.77** | **0.77** | **0.77** | **0.76** | **0.76** | **0.76** |
| Acc.  | 0.76  |        |      | 0.78  |        |      | 0.76  |        |      |

**Table 13.** Class-wise precision, recall, F1-scores, and accuracy for CoGAN, CGAN and BiGAN using RBM as the feature extractor.

|       | CoGAN |        |      | CGAN  |        |      | BiGAN |        |      |
|-------|-------|--------|------|-------|--------|------|-------|--------|------|
| $c_i$ | Prec. | Recall | F1   | Prec. | Recall | F1   | Prec. | Recall | F1   |
| 0     | 0.87  | 0.86   | 0.87 | 0.96  | 0.96   | 0.96 | 0.78  | 0.85   | 0.81 |
| 1     | 0.87  | 0.95   | 0.91 | 0.99  | 0.94   | 0.97 | 0.80  | 0.95   | 0.87 |
| 2     | 0.72  | 0.73   | 0.72 | 0.93  | 0.92   | 0.92 | 0.68  | 0.65   | 0.66 |
| 3     | **0.59** | **0.64** | **0.61** | 0.78 | 0.79 | 0.79 | 0.68 | 0.62 | 0.59 |
| 4     | 0.76  | 0.83   | 0.79 | 0.89  | 0.89   | 0.89 | 0.67  | 0.79   | 0.73 |
| 5     | **0.61** | **0.51** | **0.56** | 0.82 | 0.74 | 0.78 | **0.46** | **0.44** | **0.45** |
| 6     | 0.85  | 0.80   | 0.83 | 0.92  | 0.90   | 0.91 | 0.76  | 0.64   | 0.69 |
| 7     | 0.88  | 0.85   | 0.87 | 0.95  | 0.90   | 0.92 | 0.79  | 0.84   | 0.81 |
| 8     | 0.72  | 0.70   | 0.71 | 0.69  | 0.94   | 0.80 | **0.56** | **0.61** | **0.58** |
| 9     | 0.73  | 0.74   | 0.73 | 0.89  | 0.79   | 0.84 | 0.72  | 0.66   | 0.69 |
| Avg.  | **0.76** | **0.76** | **0.76** | **0.88** | **0.88** | **0.88** | **0.69** | **0.70** | **0.69** |
| Acc.  | 0.76  |        |      | 0.88  |        |      | 0.69  |        |      |

We shall now look at the classifier ensembles. It is important to note that the architectures of the i[th] CNN, $C_i$ and $C_i^r$ are the same. The only difference is between different $C^r$ (there is one classifier ensemble $C^r$ for each GAN) which are trained on the generated samples and tested on the original dataset. The architectural details for classifiers $C_1$ through $C_5$ (or alternatively, $C_1^r$ through $C_5^r$) are given in Table 14 through Table 18. The input shape of the generated and MNIST images were (28, 28, 3), where 28×28 were the image dimensions, and 3 was the number of channels. We also plot the validation loss and accuracy for each $C_i \in C$ over 50 epochs of training in Fig. 8. It can be inferred that as we move from $C_1$ to $C_5$, no significant change in the validation accuracy is observed. However, the validation loss at epoch 1 is reduced from $C_1$ to $C_4$ – which could be a result of increasing the number of convolutional layers, because the validation loss at epoch 1 is increased again with $C_5$ as it has only 2 convolutional layers.

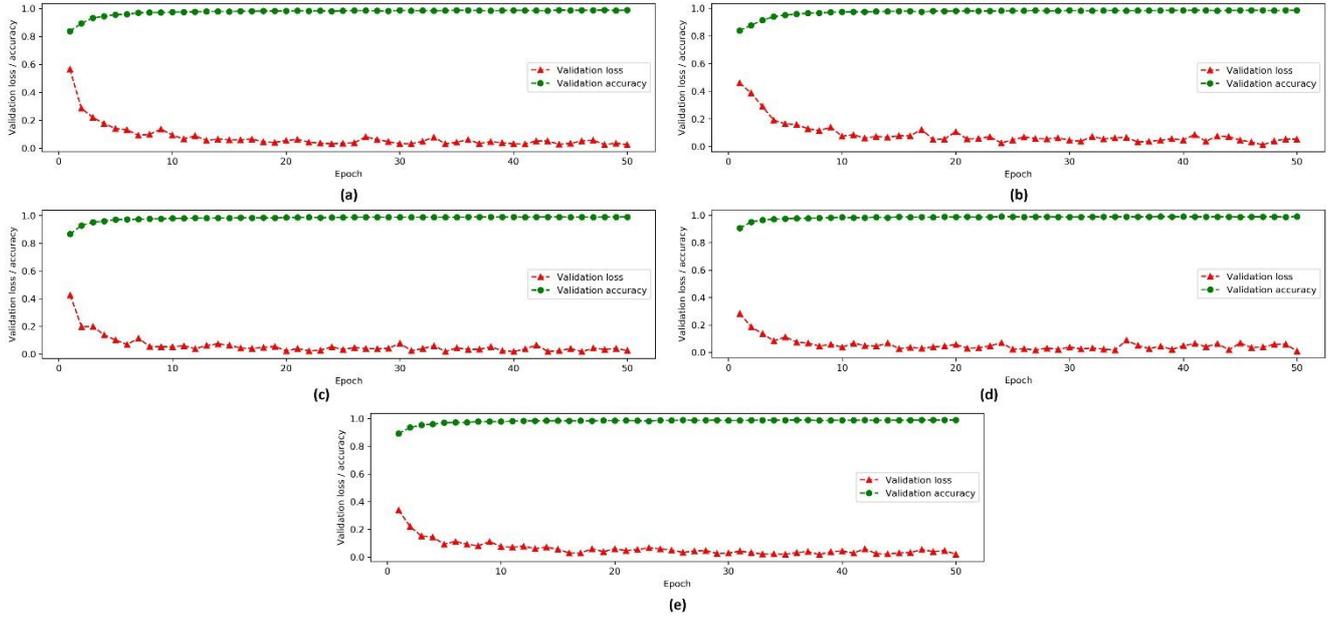

**Fig. 8.** Validation loss and accuracy for training CNNs (a) $C_1$, (b) $C_2$, (c) $C_3$, (d) $C_4$, and (e) $C_5$.

**Table 14.** Architectural details for $C_1$.

| Layer | # filters/ratio | Size/Nodes | Activation | Output shape | # params |
|---|---|---|---|---|---|
| Conv2D | 32 | (3, 3) | ReLU | (None, 26, 26, 32) | 896 |
| Conv2D | 64 | (3, 3) | ReLU | (None, 24, 24, 64) | 1.8496 |
| MaxPool2D | - | (2, 2) | - | (None, 12, 12, 64) | 0 |
| Dropout | 0.5 | - | - | (None, 12, 12, 64) | 0 |
| Flatten | - | - | - | (None, 9216) | 0 |
| Dense | - | 250 | Sigmoid | (None, 250) | $2.3 \times 10^6$ |
| Dense | - | 10 | Softmax | (None, 10) | 2510 |

**Table 15.** Architectural details for $C_2$.

| Layer | # filters/ratio | Size/Nodes | Activation | Output shape | # params |
|---|---|---|---|---|---|
| Conv2D | 32 | (3, 3) | ReLU | (None, 26, 26, 32) | 896 |
| Conv2D | 64 | (3, 3) | ReLU | (None, 24, 24, 64) | 18496 |
| MaxPool2D | - | (2, 2) | - | (None, 12, 12, 64) | 0 |
| Flatten | - | - | - | (None, 9216) | 0 |
| Dense | - | 200 | Sigmoid | (None, 200) | $1.8 \times 10^6$ |
| Dense | - | 10 | Softmax | (None, 10) | 2010 |

**Table 16.** Architectural details for $C_3$.

| Layer | # filters/ratio | Size/Nodes | Activation | Output shape | # params |
|---|---|---|---|---|---|
| Conv2D | 16 | (3, 3) | ReLU | (None, 26, 26, 16) | 448 |
| Conv2D | 32 | (3, 3) | ReLU | (None, 24, 24, 32) | 4640 |
| MaxPool2D | - | (2, 2) | - | (None, 12, 12, 32) | 0 |
| Conv2D | 64 | (3, 3) | ReLU | (None, 10, 10, 64) | 18496 |
| Dropout | 0.3 | - | - | (None, 10, 10, 64) | 0 |
| Flatten | - | - | - | (None, 6400) | 0 |
| Dense | - | 300 | Sigmoid | (None, 300) | $1.9 \times 10^6$ |
| Dense | - | 10 | Softmax | (None, 10) | 3010 |

**Table 17.** Architectural details for $C_4$.

| Layer | # filters/ratio | Size/Nodes | Activation | Output shape | # params |
|---|---|---|---|---|---|

| Layer | # filters/ratio | Size/Nodes | Activation | Output shape | # params |
|---|---|---|---|---|---|
| Conv2D | 16 | (3, 3) | ReLU | (None, 26, 26, 16) | 448 |
| Conv2D | 32 | (3, 3) | ReLU | (None, 24, 24, 32) | 4640 |
| MaxPool2D | - | (2, 2) | - | (None, 12, 12, 32) | 0 |
| Conv2D | 64 | (3, 3) | ReLU | (None, 10, 10, 64) | 18496 |
| Flatten | - | - | - | (None, 6400) | 0 |
| Dense | - | 225 | Sigmoid | (None, 225) | $1.4 \times 10^6$ |
| Dense | - | 10 | Softmax | (None, 10) | 2260 |

**Table 18.** Architectural details for $C_5$.

| Layer | # filters/ratio | Size/Nodes | Activation | Output shape | # params |
|---|---|---|---|---|---|
| Conv2D | 32 | (3, 3) | ReLU | (None, 26, 26, 32) | 896 |
| MaxPool2D | - | (2, 2) | - | (None, 13, 13, 32) | 0 |
| Conv2D | 64 | (3, 3) | ReLU | (None, 11, 11, 64) | 18496 |
| Dropout | 0.3 | - | - | (None, 11, 11, 64) | 0 |
| Flatten | - | - | - | (None, 7744) | 0 |
| Dense | - | 225 | Sigmoid | (None, 225) | $1.7 \times 10^6$ |
| Dense | - | 10 | Softmax | (None, 10) | 2260 |

Now, we calculate the GM Score $\mathcal{G}$ for each GAN based on the results from Table 3, Table 4, and Table 8 through Table 13. The proposed metric is calculated for each GAN and provided in Table 19. It can be seen that BiGAN scores the highest ($\mathcal{G} = 0.5065$) because of adequate inter-class diversity and a very high intra-class diversity. From Table 19, it is important to note that every model had more or less the same ES (barring CGAN), which may be attributed to the fact that $C_1$ was used to separate the generated images for all models except CGAN – this step was done before calculating ES as training CNNs falls under supervised learning, and for the same, image labels are needed. Therefore, the images that $C_1$ classified were tested again by classifiers in ensemble $C$ and $C^r$ which are not drastically different in their architectures, hence, every ensemble score is more or less the same. This also explains why CGAN has a different ES than other models – because $C_1$ was not used to separate the images into respective classes (i.e. obtain labels for generated images). When the number of generated samples are lower, we advise that the assignment of class labels to samples be done manually, so that ES may differ from model to model. One may get an idea of the sample fidelity by looking at the DBM, RBM, and ES scores. In our case, image fidelity can be understood through the DBN and RBM average of all metrics (precision, recall, F1-score, and accuracy). Also, while CGAN has a perfect inter-class diversity (which should not be evaluated as the model basically cheats by conditional sampling from a known distribution), it still does not have the highest $\mathcal{G}$ owing to a very low intra-class diversity score. We shall discuss more about these results in Section 7.

**Table 19.** Final score calculation, where DBN+RBM (avg) refers to the average of metrics of precision, recall, F1-score and accuracy for both RBM and DBM-based approaches.

| Model | $\mathbb{D}^i$ | $\mathbb{D}^{\cdots}$ | DBN+RBM (avg) | ES | $\mathcal{G}$ |
|---|---|---|---|---|---|
| WGAN | 0.7024 | 0.2436 | 0.5399 | 0.9997 | 0.1848 |
| WGAN_I | 0.7942 | 0.3078 | 0.7712 | 0.9997 | 0.3770 |
| SGAN | 0.9017 | 0.2846 | 0.7825 | 0.9997 | 0.4015 |
| LSGAN | 0.8270 | 0.3106 | 0.7924 | 0.9996 | 0.4069 |
| GAN | 0.8120 | 0.2763 | 0.8075 | 0.9996 | 0.3622 |
| DCGAN | **0.9334** | 0.2795 | 0.7949 | 0.9997 | 0.4147 |
| CoGAN | 0.7447 | 0.3195 | 0.7937 | 0.9996 | 0.3777 |
| CGAN | 1.0000 | 0.1500 | **0.9000** | 0.9990 | 0.2699 |
| BiGAN | 0.8181 | **0.4263** | 0.7262 | 0.9996 | **0.5065** |

## 7. Discussion

In this section, we shall discuss how our metric compares with those in the literature and why the proposed metric is better, how our calculation of intra-class diversity is intuitive (i.e. model captures more modes of the training data), how image fidelity scores are directly related to the quality of the samples, among other things. As we saw in Section 3, there have been many attempts at the evaluation of GANs, with there being no consensus on the best GAN evaluation measure. The Inception Score (IS) has been adopted widely but has its own shortcomings, e.g. it does not consider intra-class diversity. To summarize, we compare other metrics with ours in Table 20.

It can be seen from Table 20. that no metric compares exactly with the proposed GM Score. It would seem that many metrics suffer from not calculating the intra-class diversity, which is generally more difficult than calculating inter-class diversity. Moreover, the metrics that do consider intra-class diversity, i.e. m-IS, FID, and Wasserstein Critic have other drawbacks such as high computational complexity for metric calculation, assumption of InceptionNet layer embedding to be a continuous multivariate Gaussian, and no consideration for inter-class diversity, respectively.

**Table 20.** Comparison of different GAN evaluation metrics, including proposed metric. 'High', 'Moderate' and 'Low' are scores given to models where a numeric entry is not available. 'Conformity' refers to how well the score conforms to human perception, 'Fidelity' refers to the consideration of sample quality, and finally, 'N/A' refers to there being insufficient data/availability of varied data.

| Metric | Ref. | Bounds | Conformity | Latent space | Intra-class diversity | Inter-class diversity | Fidelity |
|---|---|---|---|---|---|---|---|
| Inception Score (IS) | [23] | $[1, \infty]$ | High | ✗ | ✗ | ✓ | ✓ |
| Modified-IS (m-IS) | [26] | $[1, \infty]$ | High | ✗ | ✓ | ✓ | ✓ |
| Fréchet Inception Distance (FID) | [28] | $[0, \infty]$ | High | ✗ | ✓ | ✓ | ✓ |
| Generative Adversarial Metric (GAM) | [31] | N/A | High | ✗ | ✗ | ✗ | ✓ |
| GAN Quality Index (GQI) | [32] | $[0, 100]$ | High | ✗ | ✗ | ✗ | ✓ |
| Average Log-likelihood | [1] | $[-\infty, \infty]$ | Low | ✗ | ✗ | ✗ | ✗ |
| AM Score | [40] | $[0, \infty]$ | High | ✗ | ✗ | ✓ | ✓ |
| Coverage Metric | [41] | $[0, 1]$ | Low | ✗ | ✗ | ✗ | ✗ |
| Mode Score (MS) | [42] | $[0, \infty]$ | High | ✗ | N/A | ✓ | ✓ |
| Maximum Mean Discrepancy (MMD) | [43] | $[0, \infty]$ | N/A | ✗ | ✗ | N/A | N/A |
| Wasserstein Critic | [13] | $[0, \infty]$ | N/A | ✗ | ✓ | ✗ | ✓ |
| **GM Score** | | **[0, 1]** | **High** | ✓ | ✓ | ✓ | ✓ |

In Section 4.2, Remark 1, we had mentioned how WGAN had a very high intra-class diversity and hence the overdiversity coefficient was applied for regularization. As comparison, we show in Fig. 9. how WGAN's generation relates directly to how high its actual, unregulated mean entropy was and its DBN and RBM average metric scores, in comparison to samples generated by SGAN. It is evident that the overdiversity coefficient $\beta$ is very important so that models do not exploit by attaining higher entropy through bad quality of samples.

As far as intra-class diversity is concerned, there is one key takeaway from Table 4. The standard deviations $\sigma$ play an important role in conveying how the class-wise entropy distributions are spread about their mean, i.e. whether they are sharp or blunt. As discussed in Section 4.2, these values can help in the detection of mode collapse of the generator. Moreover, a high $\sigma$ indicates that the entropy mass has a higher spread about the mean, conveying that the samples in $I_g$ vary in terms of their visual features, which is desirable for a generator. The incorporation of $\sigma$ could have been possible, but because there is already a positive correlation between $\sigma$ and $\mathbb{D}^{\cdots}(G(\mathbf{z}))$, doing so would prove to be unnecessary.

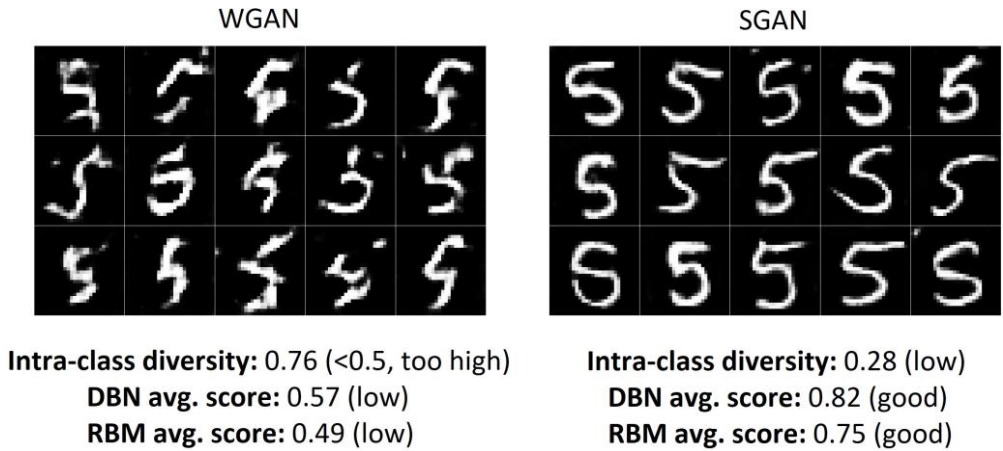

**Fig. 9.** While WGAN may be attempting to display more modes of data relative to SGAN, it suffers from low sample fidelity, for which the DBN and RBM average metric scores are low. Because the intra-class diversity is so high ($> \beta$), it is regularized

to become 0.24 according to (5). On the other hand, in spite of having good DBN+RBM average, SGAN suffers from lack of variety in generated samples, hence the lower intra-class diversity.

In Remark 3, we have already discussed how the proposed metric stands biased for CGAN due to its conditional sampling from a known distribution. This gives CGAN an unfair advantage both in terms of inter-class diversity (which would be unity, i.e. perfect), and a higher average score for evaluation metrics of DBN and RBM because of familiarity of true data modes, as discussed in Remark 3. In spite of these, CGAN suffers from a lack of intra-class diversity, achieving a very low score of $\mathbb{D}^{\cdots}(G(z)) = 0.15$. This is explainable behaviour; as CGAN is given conditional sampling, it becomes easy for the model to fall into capturing only a few modes by generating similar samples from a given conditioning. As long as the discriminator is fooled, the generator continues generating images which may be visually similar. This is not necessarily a drawback for the proposed metric, as it is illogical to evaluate the class diversity of generative models that take prior information of the class to generate.

We argue that the proposed metric is applicable to data beyond the domain of images, too. The class-wise entropies are obtained from trained CNNs which operate well over images, but a similar procedure may be conducted for, say, time-series data with different architectures of recurrent neural networks (RNN) [47]. Similarly, the ES score can also be obtained. RBM and DBN are known feature extractors which work irrespective of the domain of the data. This proves that the proposed metric is independent of the domain of the data, which is image in our case. Also, the reader may be inclined to conclude that the proposed score is only for the evaluation of GANs, however, the score may be applied to any generative model such as the DBN, RBM, VAE, etc. that are employed for generating data.

The calculation of inter-class diversity $\mathbb{D}^{:}$ depends on whether the training data is evenly distributed (by even, we mean that there should not be an extreme class imbalance). It is important, while calculating GM Score, that the training classes are evenly distributed so that GANs have equal opportunity to generate every class, i.e. there must be no bias in the generated class diversity. If there is a high imbalance, it would not be convenient to calculate the generated sample counts' distance from the mean, as we do in $\mathbb{D}^{:}$. In such a scenario, it would better to minimize the KL divergence between the generated class count distribution and the training class count distribution. The AM Score addresses this issue in a similar fashion.

It is also important to note the reason why we only chose MNIST as the dataset apart from other popular datasets like CIFAR-10 [44], CelebA [45], etc. The motive of our work in this paper was to introduce a new GAN evaluation metric that would overcome the shortcomings of other metrics. It is trivial to question whether the proposed metric is applicable to other datasets – because the metric depends on the generated samples, and not the base dataset itself, i.e. it is independent of the domain of the training data. Thus, we decided not to go with other datasets because of the triviality involved, along with other reasons such as hardware limitations, the increased time of training all the GANs on new datasets, and the domain independence of the proposed metric. In the future, we may explore how the GM Score varies for the same set of GANs from dataset to dataset.

## 8. Conclusion

In this paper, we propose a new evaluation metric for generative models, namely, the GM Score. The following GANs are compared: Wasserstein GAN, Wasserstein GAN improved, semi-supervised GAN, least-squares GAN, vanilla GAN, deep convolutional GAN, coupled GAN, conditional GAN, and bidirectional GAN. These models are trained over the MNIST dataset for image generation. The proposed metric successfully answers whether the generated samples are (a) diverse enough in terms of their class, (b) diverse enough in terms of visual variety in each individual class, and (c) if samples are of high quality or fidelity through two discriminability tests. These discriminability tests involve training of classifier ensembles and powerful feature extractors such as the deep belief network (DBN) and the restricted Boltzmann machine (RBM). We compare the proposed technique with other existing techniques and show how our technique overcomes the shortcomings of other approaches in the literature.